\documentclass[11pt]{article}

\usepackage[final]{acl}
\usepackage{times}
\usepackage{latexsym}
\usepackage[T1]{fontenc}
\usepackage[utf8]{inputenc}
\usepackage{microtype}
\usepackage{inconsolata}
\usepackage{tcolorbox}
\usepackage{graphicx}
\usepackage{multirow}
\usepackage{booktabs}
\usepackage{hhline}
\usepackage{enumitem}
\usepackage{amsmath}
\usepackage{amssymb}
\usepackage{algorithm}
\usepackage{algorithmic}
\usepackage{xcolor}
\usepackage[table]{xcolor}
\usepackage{listings}

\title{From Tags to Trees: Structuring Fine-Grained Knowledge for Controllable Data Selection in LLM Instruction Tuning}

\newcommand*{\affaddr}[1]{#1}
\newcommand*{\affmark}[1][*]
{\textsuperscript{#1}}
\newcommand*{\email}[1]{\texttt{#1}}

\author{
Zihan Niu\affmark[\textnormal{1}]\thanks{Work done during internship at Kuaishou.}, 
Wenping Hu\affmark[\textnormal{2}]\footnotemark[2], 
Junmin Chen\affmark[\textnormal{2}],
Xiyue Wang\affmark[\textnormal{2}]\footnotemark[1],  \\
\textbf{Tong Xu}\affmark[\textnormal{1}]\thanks{Corresponding authors.},
\textbf{Ruiming Tang}\affmark[\textnormal{2}]\\
\affaddr{\affmark[1]University of Science and Technology of China} \\
\affaddr{\affmark[2]Klear Team, Kuaishou Technology} \\
\email{niuzihan@mail.ustc.edu.cn, tongxu@ustc.edu.cn} \\
\email{\{huwenping, chenjunmin, wangxiyue, tangruiming\}@kuaishou.com}
}

\begin{document}
\maketitle
\begin{abstract}
Effective and controllable data selection is critical for LLM instruction tuning, especially with massive open-source datasets. 
Existing approaches primarily rely on instance-level quality scores, or diversity metrics based on embedding clusters or semantic tags.
However, constrained by the flatness of embedding spaces or the coarseness of tags, these approaches overlook fine-grained knowledge and its intrinsic hierarchical dependencies, consequently hindering precise data valuation and knowledge-aligned sampling.
To address this challenge, we propose Tree-aware Aligned Global Sampling (TAGS), a unified framework that leverages a knowledge tree built from fine-grained tags, thereby enabling joint control of global quality, diversity, and target alignment.
Using an LLM-based tagger, we extract atomic knowledge concepts, which are organized into a global tree through bottom-up hierarchical clustering. 
By grounding data instances onto this tree, a tree-aware metric then quantifies data quality and diversity, facilitating effective sampling. Our controllable sampling strategy maximizes tree-level information gain and enforces leaf-level alignment via KL-divergence for specific domains. 
Extensive experiments demonstrate that TAGS significantly outperforms state-of-the-art baselines. Notably, it surpasses the full-dataset model by \textbf{+5.84\%} using only \textbf{5\%} of the data, while our aligned sampling strategy further boosts average performance by \textbf{+4.24\%}.
\end{abstract}

\section{Introduction}
Large Language Models (LLMs) have become a cornerstone of natural language processing, demonstrating remarkable performance across diverse applications~\cite{yang2025qwen3, liu2024deepseek, achiam2023gpt}. While pre-training endows LLMs with extensive world knowledge, instruction tuning serves to bridge the gap between raw probabilistic modeling and human intent~\cite{ouyang2022training, wei2021finetuned}. As a result, the quality and diversity of instruction data have become critical factors in determining the performance of LLMs.
\begin{figure}
    \centering
    \includegraphics[width=0.9\linewidth]{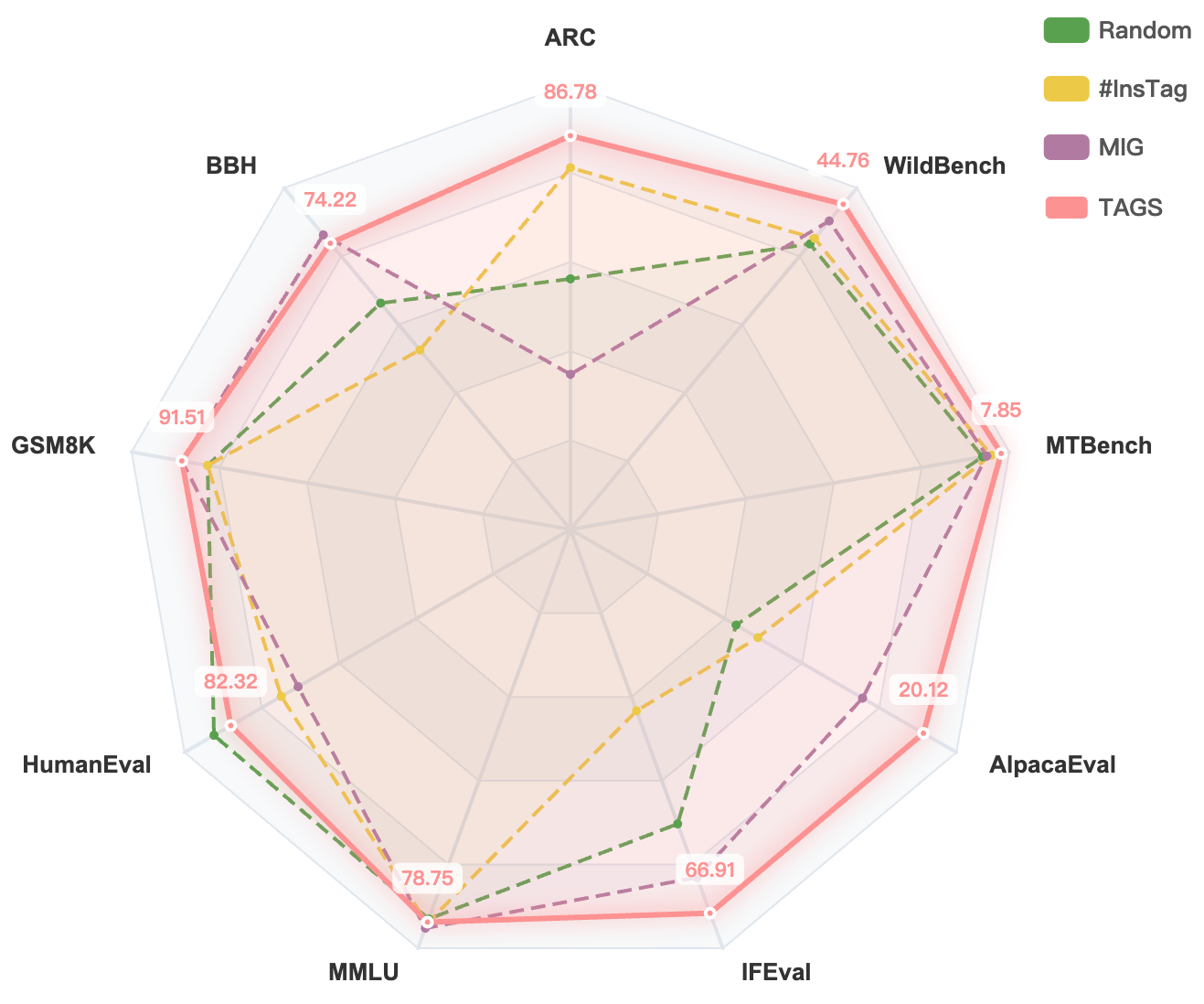}
    \caption{Performance comparison of Qwen3-8B fine-tuned on Tulu3. TAGS consistently outperforms other data selection strategies.}
    \label{fig:comparison}
    \vspace{-1.5em}
\end{figure}
The recent emergence of large-scale open-source instruction datasets~\cite{lambert2024tulu, OpenHermes} has significantly democratized access to instruction tuning. However, simply scaling up data size does not always guarantee better performance. Recent studies~\cite{zhou2023lima, ge2024clustering, chen2025mig} reveal the \textit{"less is more"} phenomenon, where models trained on carefully curated, high-quality subsets consistently outperform those trained on massive, unfiltered corpora. These findings have shifted research attention from indiscriminate scaling toward effective data selection. In particular, extracting high-value instruction subsets not only reduces training costs but also mitigates the \textit{alignment tax}—the degradation in model capability caused by redundant, noisy instructions~\cite{ouyang2022training, zhou2023lima}. As such, effective data selection has become critical in ensuring optimal LLM performance.

Driven by this demand, numerous automated data selection methods have been proposed~\cite{lu2023instag, ge2024clustering, chen2025mig}. These approaches can generally be categorized into three types: metric-based filtering, diversity-driven selection, and comprehensive strategies.
Metric-based filtering methods~\cite{chen2023alpagasus, li2024quantity, liu2024selectit} rely on scalar indicators such as quality or complexity scores to filter out noisy data. However, they evaluate samples independently, failing to consider global diversity at the subset level.
Diversity-driven selection methods~\cite{yin2024entropy, bukharin2024data} focus on maximizing semantic coverage. Nevertheless, the use of flattened representations limits their ability to capture fine-grained conceptual distinctions.
Comprehensive strategies~\cite{lu2023instag, ge2024clustering, chen2025mig} attempt to balance quality and diversity by using tagging or clustering. Despite this, these methods typically operate over coarse labels (e.g., \textit{“Math”}) and shallow structures, which cannot distinguish atomic concepts like \textit{"matrix decomposition"} and \textit{"linear equations"}.
Crucially, all these paradigms operate over flat semantic spaces and overlook fine-grained knowledge and its intrinsic tree-structured hierarchy.
This leads to two fundamental bottlenecks:
(1) Imprecise assessment of data value and subset diversity due to the lack of a fine-grained topology.
(2) Uncontrollable sampling that struggles to meet specific domain constraints or user requirements.

To address these limitations, we propose TAGS (\textbf{T}ree-aware \textbf{A}ligned \textbf{G}lobal \textbf{S}ampling), a unified framework that incorporates fine-grained knowledge topology into controllable data selection.
First, we construct a global Knowledge Tree to capture fine-grained knowledge structures. Rather than relying on coarse labels, we utilize a specialized lightweight tagger to extract atomic concepts from large-scale data~\cite{xu2024magpie, numina_math_datasets, zheng2023lmsyschat1m, OpenHermes}. These concepts are then organized via hierarchical clustering, forming a tree where leaf nodes represent specific knowledge units and upper levels represent broader semantic categories.
Second, we introduce a Tree-based Information Propagation mechanism to evaluate data utility in a structured semantic space.
Specifically, TAGS anchors data instances onto the constructed tree, evaluating data value through a tree-based information flow.
Data quality drives the information intensity at leaf nodes, which then propagates to quantify global richness of the subset.
To mitigate redundancy, we employ a concave aggregation function that ensures diminishing returns for repeated concepts, thus balancing local quality with global diversity.
Finally, we implement an efficient data selection algorithm with two controllable strategies: maximizing global information gain for \textbf{General Sampling}, and incorporating KL-divergence for \textbf{Knowledge-Aligned Sampling} to approximate target domain distributions.

To validate the effectiveness of TAGS, we conducted extensive experiments using the Qwen3~\cite{yang2025qwen3} model family at varying scales, across a diverse set of knowledge-intensive~\cite{clark2018think, suzgun2023challenging, hendrycks2020measuring, chen2021evaluating, cobbe2021training, zhou2023instruction} and human-preference benchmarks~\cite{zheng2023judging, dong2024abilities, lin2024wildbench}.
As shown in Figure~\ref{fig:comparison}, TAGS consistently achieves state-of-the-art performance, significantly outperforming strong baselines in both general capabilities and reasoning tasks. Remarkably, using only \textbf{5\%} of the data, TAGS outperforms the full-dataset fine-tuning baseline by \textbf{2.23\%} on knowledge-intensive benchmarks and achieves a substantial \textbf{12.34\%} gain in human-preference evaluations.
Moreover, TAGS demonstrates precise controllability through knowledge-aligned sampling, with an average performance improvement of \textbf{4.24\%} on target domains compared to the general selection strategy, highlighting the efficacy of domain-specific adaptation.

In summary, our contributions are as follows:
\begin{itemize}[leftmargin=*, nosep]
\item We propose TAGS, a unified framework that leverages hierarchical structures built from fine-grained tags to enable global quality, diversity, and target alignment in data selection.
\item We introduce two sampling strategies, General Sampling and Knowledge-Aligned Sampling, offering flexible and controllable data selection to meet diverse requirements.
\item Extensive evaluations across model scales demonstrate state-of-the-art data efficiency.
\end{itemize}

\section{Related Work}
\begin{figure*}
    \centering
    \includegraphics[width=1.0\linewidth]{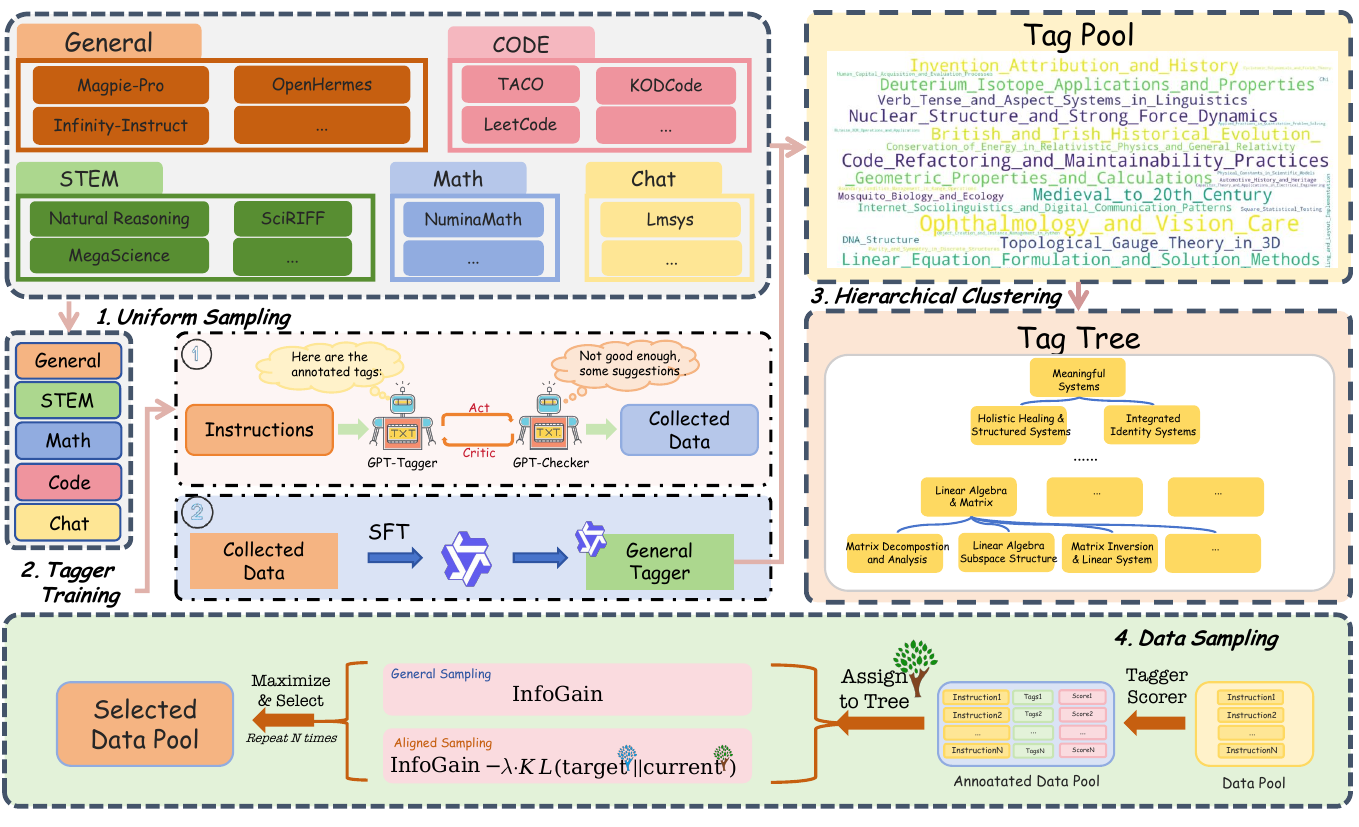}
    \caption{Overview of the TAGS framework: (1) \textbf{Uniform Sampling} from diverse data sources; (2) \textbf{Tagger Training}, which employs an LLM-based Act-Critic loop to synthesize fine-grained tags and fine-tune a General Tagger; (3) \textbf{Hierarchical Clustering} structures the generated tag pool into a global Tag Tree; and (4) \textbf{Data Sampling} maps instances to tree nodes and executes controllable selection based on Information Gain and KL-divergence.}
    \label{fig:overview}
    \vspace{-1.0em}
\end{figure*}

\textbf{Instruction Tuning and Data Efficiency.} Instruction tuning is widely acknowledged as an important approach for aligning LLMs with human intent ~\cite{ouyang2022training, wei2021finetuned, sanh2021multitask}. 
Recent studies~\cite{zhou2023lima, ge2024clustering,chen2025mig} further demonstrate that during the Supervised Fine-Tuning (SFT) stage, utilizing a meticulously curated subset of high-quality instructions can yield performance superior to models trained on significantly larger datasets. 
These findings underscore the importance of strategic data selection, where developing automated frameworks to balance both data quality and diversity has emerged as a vital research topic.

\noindent    \textbf{Automated Data Selection.} 
Existing automated data selection methods~\cite{chen2023alpagasus,lu2023instag,ge2024clustering,liu2024selectit,chen2025mig} aim to identify training subsets that maximize efficiency and performance. Broadly, they can be categorized into metric-based filtering, diversity-driven selection, and comprehensive strategies.
(1) \textit{\textbf{Metric-based filtering}} relies on scalar criteria to assess data quality or difficulty. IFD~\cite{li2024quantity} estimates instruction difficulty via the discrepancy between predicted and self-generated responses, while ALPAGASUS ~\cite{chen2023alpagasus} employs LLM-based judges and SelectIT ~\cite{liu2024selectit} filters samples based on intrinsic uncertainty. Although effective at denoising, these methods evaluate instances independently, ignoring the global diversity.
(2) \textit{\textbf{Diversity-driven selection}} emphasizes coverage of the semantic space. ZIP ~\cite{yin2024entropy} promotes diversity through a compression-based metric, whereas QDIT~\cite{bukharin2024data} selects representative samples that maximize current subset diversity.
(3) \textit{\textbf{Comprehensive strategies}} seek to jointly balance quality and diversity during the selection. CaR~\cite{ge2024clustering} performs clustering followed by intra-cluster quality selection, while DEITA~\cite{liumakes} greedily selects high-quality samples with semantic redundancy penalties. \textit{\#InsTag}~\cite{lu2023instag} introduces LLM-generated coarse-grained tags and selects data based on tag coverage and complexity, MIG ~\cite{chen2025mig} further extends this by modeling semantic relations as a label graph and maximizing information gain.
Despite these advances, existing methods overlook fine-grained knowledge and their explicit hierarchical dependencies, limiting the capability for precise data metric and knowledge-aligned selection.
To address this issue, we introduce TAGS, which leverages fine-grained tags and tree structure to achieve global quality, diversity, and targeted sampling.

\section{Method}
As shown in Figure~\ref{fig:overview}, we propose TAGS, a tree-aware data selection framework that enables controllable and knowledge-aligned sampling.
TAGS fine-tunes an LLM to extract fine-grained tags (Sec.~\ref{sec:tagger}), organizes them into a Tag Tree (Sec.~\ref{sec:tree}), and performs data selection via Information Gain and KL-divergence (Sec.~\ref{sec:sampling}).

\subsection{Problem Formulation \& Notation}
\textbf{Task.} 
Given a dataset $D_P$, a budget $N$ and a large language model $M$, the task is to select a subset $D_S \subset D_P$ of size $N$ such that the performance of the LLM fine-tuned on $D_S$, denoted by $F(M, D_S)$, is maximized. This can be formulated as:
\begin{equation}
D_S = \operatorname*{argmax}_{D \subset D_P, \, |D|=N} F(M, D).
\end{equation}

\noindent \textbf{Data.} 
Each data instance $d_i$ consists of a query-response pair $(q_i, r_i)$, along with associated attributes $(T_i, Q_i, C_i)$, where $q_i$ denotes the query and $r_i$ denotes the response. 
$T_i$ represents extracted fine-grained tags, while $Q_i$ and $C_i$ denote the quality score and complexity score, respectively, which are separately evaluated by the two scorers introduced in DEITA~\cite{liumakes}.

\subsection{Fine-grained Tagger Training}
\label{sec:tagger}
\begin{figure}
    \centering
    \includegraphics[width=0.95\linewidth]{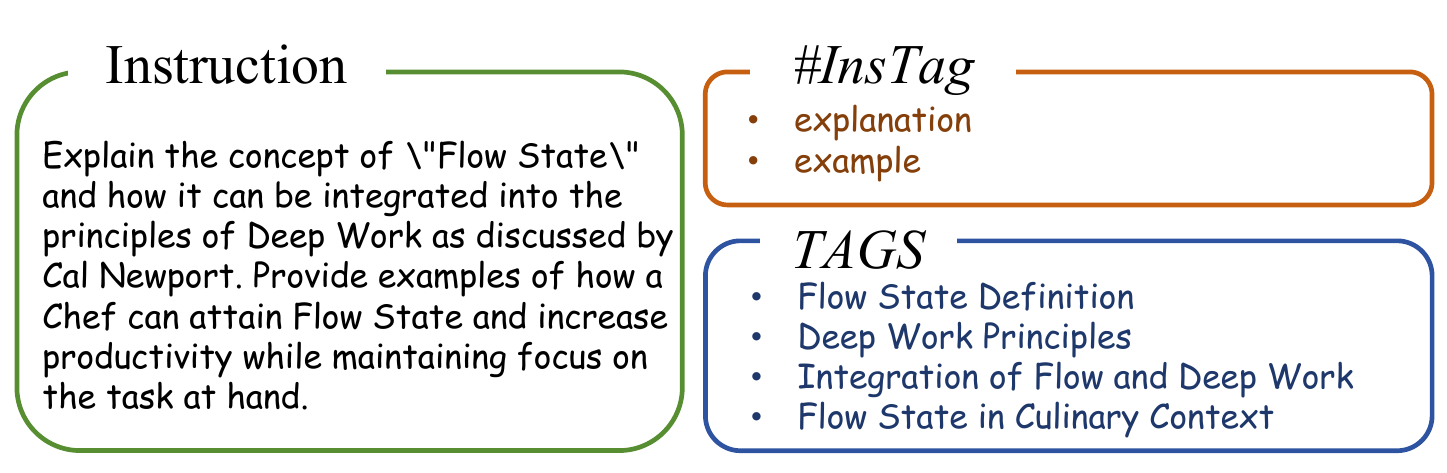}
    \caption{Comparison of \textit{\#InsTag} and TAGS tagger.}
    \label{fig:tagger}
    \vspace{-1.4em} 
\end{figure}

When dealing with massive instructions, existing approaches~\cite{ge2024clustering,bukharin2024data} typically rely on embeddings and clustering to evaluate data distributions and diversity. 
However, these methods lack interpretability, making it difficult to distinguish whether clusters are driven by shared linguistic style, core knowledge, or other latent factors.
In contrast, explicit semantic tags map unstructured text into lower-noise tag spaces that better capture the underlying knowledge, enabling data to be measured in a more compact and controllable semantic space. 
Nevertheless, as illustrated in Figure~\ref{fig:tagger}, existing taggers (e.g., \textit{\#InsTag}) primarily focus on coarse-grained intents and struggle to model the specific entities, contexts, and constraints embedded in complex instructions.
This limitation motivates the development of a more precise, fine-grained tagger, which enables reliable extraction of fine-grained knowledge in TAGS.

We construct a high-quality training corpus for fine-grained tagging, consisting of 50k samples spanning core domains, augmented with an additional 3k manually curated complicated instructions to enhance the tagger’s capability in handling complex queries (see Table~\ref{tab:data_domain} for a detailed breakdown).
Inspired by the Self-Refine mechanism \cite{madaan2023self}, we implement an Act-Critic annotation pipeline powered by \texttt{GPT-o4-mini}. In this pipeline, a \textit{GPT-Tagger} generates initial tags, while a \textit{GPT-Checker} reviews them and provides refinement suggestions. 
To balance quality and efficiency, we limit the annotation process to a maximum of three iterations.
Finally, we perform SFT on \texttt{Qwen3-8B-Instruct}~\cite{yang2025qwen3} using this synthetic data, yielding a general tagger that balances precise query understanding with inference efficiency for large-scale data processing.

\subsection{Hierarchical Tag Tree Construction}
\label{sec:tree}
Despite the semantic denoising achieved by TAGS tagger, the resulting tag pool still exhibits significant fragmentation.
Similar concepts may be assigned different tags due to variations in expression (e.g., \textit{"Value-to-Index Mapping"} vs. \textit{"Index-based count computation"}), while flat tag representations struggle to reflect the hierarchical structure of complex instructions (e.g., \textit{"Matrix Decomposition"} under \textit{"Linear Algebra"}).
As a result, isolated tags limit the global assessment of data coverage, motivating the need for a structured representation of tag semantics.
Inspired by Clio~\cite{tamkin2024clio}, we construct a Tag Tree for downstream analysis and sampling.
To balance efficiency with precision, we adopt an iterative bottom-up framework. 
At each iteration, we first aggregate the current set of nodes into clusters using K-Means based on semantic embeddings\footnote{We utilize \texttt{Qwen3-Embedding-0.6B} as encoder.}.
To resolve ambiguities inherent in embedding-based clustering, we employ an LLM\footnote{\texttt{Qwen3-30B-A3B-2507}} to refine these clusters through a structured four-step pipeline:
\begin{itemize}[leftmargin=*, nosep]
\item \textbf{Summarization:} Abstracting a representative topic candidate for each cluster;
\item \textbf{Deduplication:} Removing synonymous topics to ensure uniqueness;
\item \textbf{Reassign:} Reviewing member tags and reassigning misclassified items to their appropriate topic to rectify vector-based errors;
\item \textbf{Renaming:} Refining the topic name to accurately reflect the corrected cluster composition.
\end{itemize}
The generated topics serve as inputs for the subsequent iteration. This cycle repeats recursively until a specified tree depth is reached, ultimately constructing the global Tag Tree.

\subsection{Tree-aware Data Sampling}
\label{sec:sampling}
\noindent
\textbf{Tree Topology \& Data Anchoring.}
We utilize the semantic tag tree $\mathcal{T} = (V, E)$ constructed in Sec.~\ref{sec:tree}, where $V$ denotes the set of nodes and $E$ represents the directed edges encoding hierarchical relations.
To facilitate data anchoring, we transform this topology into an ancestry matrix $\mathbf{M} \in \{0,1\}^{|V| \times |V_{\text{leaf}}|}$.
This matrix maps the relationships between all nodes $V$ and the subset of fine-grained leaf nodes $V_{\text{leaf}} \subset V$:
\begin{equation}
\label{eq:ancestor_matrix}
\mathbf{M}_{ij} = \mathbb{I}(i \text{ is an ancestor of } j) \lor \mathbb{I}(i = j),
\end{equation}
where $\mathbb{I}(\cdot)$ is the indicator function.
This explicitly encodes the logical implication: if a leaf concept is active, all its ancestral categories are logically implicated.
We further anchor the training instances to this topology to obtain a structure-aware representation.
Each sample $d_k$ is assigned to leaf nodes by matching its fine-grained tags $T_k$ to the leaf-level concepts based on maximum semantic similarity, resulting in a binary leaf activation vector
$\mathbf{h}^{(\text{leaf})}_k \in \{0,1\}^{|V_{\text{leaf}}|}$.
To reconstruct the tree-level semantic profile, we propagate leaf-level signals upward via the ancestry matrix:
\begin{equation}
\label{eq:tree-activation}
\mathbf{h}^{(\text{tree})}_k = \mathbf{M} \cdot \mathbf{h}^{(\text{leaf})}_k.
\end{equation}

\noindent
\textbf{Tree-based Information Modeling.}
Given the hierarchical profile \(\mathbf{h}^{(\text{tree})}_k\) of each data point \(d_k\), we first quantify its utility by combining its quality score \(Q_k\) and complexity score \(C_k\).
We define a composite scalar weight \(s_k\) as:
\begin{equation}
s_k = \alpha \cdot Q_k + (1-\alpha) \cdot C_k,
\end{equation}
where \(\alpha \in [0,1]\) is a hyperparameter balancing the two metrics.
To further model the effective contribution of \(d_k\) to the semantic space, we scale its profile by $s_k$, yielding the raw information vector:
\begin{equation}
\mathbf{e}_k = s_k \cdot \mathbf{h}^{(\text{tree})}_k .
\end{equation}

Since concepts are organized along a hierarchy, treating them as isolated features fails to capture their semantic dependencies.
To capture this semantic overlap, we allow information to propagate through the tag tree structure.
Following~\citet{chen2025mig}, we define the propagation coefficient $\mathbf{A}_{pq}$ from node $p$ to node $q$ based on the tree topology:
\begin{equation}
\label{eq:prop}
    \mathbf{A}_{pq} = \frac{a_{pq}}{1+\sum_{j \neq p} a_{pj}},
\end{equation}
where $a_{pq} = \mathbb{I}((p,q) \in E)$ indicates the binary structural connection strength.
This propagation mechanism ensures that effective coverage is measured across related conceptual branches rather than discrete nodes.

Based on this propagation matrix $\mathbf{A}$, the effective information vector of $d_k$ is transformed into $\mathbf{A}\mathbf{e}_k$.
Finally, we quantify the utility of the selected subset \(D\) by aggregating the propagated information:
\begin{equation}
\mathcal{I}(D) = \sum_{j \in V} \Phi (\sum_{k \in D} A\mathbf{e}_k)
\end{equation}
where \(\Phi(x) = x^\gamma\) (with \(\gamma \in (0,1)\)) is a strictly increasing, concave function.
Crucially, $\Phi$ being increasing prioritizes samples with high composite scores, while its concavity imposes diminishing returns to mitigate redundancy, thereby balancing individual quality and global diversity.

\noindent
\textbf{Tree-aware Dual-mode Sampling.}
To curate a high-quality subset, we adopt a greedy strategy based on Information Gain.
However, re-evaluating the full objective for every candidate is computationally expensive.
To ensure efficiency, we approximate the marginal information gain using a first-order Taylor expansion:
\begin{equation}
\label{eq:gain}
\small
\begin{aligned}
d_k 
&= \operatorname*{argmax}_{d \in D_P^k} \left( \mathcal{I}(D_S^k \cup \{d\}) - \mathcal{I}(D_S^k) \right) \\
&\approx \operatorname*{argmax}_{d \in D_P^k} \left( \nabla_{\mathbf{v}} \mathcal{I}(\mathbf{v}_k)^\top (\mathbf{A} \mathbf{e}_d) \right) \\
&= \operatorname*{argmax}_{d \in D_P^k} ( \Phi' ( \mathbf{A} \sum_{i \in D_S^k} \mathbf{e}_i )^\top \mathbf{A} \mathbf{e}_d ),
\end{aligned}
\end{equation}
where $\mathbf{v}_k = \mathbf{A} \sum_{i \in D_S^k} \mathbf{e}_i$ denotes the accumulated semantic profile, while $D_S^k$ and $D_P^k$ represent the selected subset and the remaining candidate pool at iteration $k$, respectively.

To accommodate diverse application needs, we extend this framework with two distinct sampling modes:
(1) \textit{General Sampling}, which prioritizes global quality and diversity across the entire Tag Tree; and 
(2) \textit{Aligned Sampling}, which enforces adaptation to specific domains.
For the latter, we introduce a target distribution $Q \in \Delta^{|V_{\text{leaf}}|}$ defined over the leaf nodes.
Since leaves represent atomic knowledge units, $Q$ allows for fine-grained control over the desired dataset composition, whether specified manually for domain customization or derived from a reference benchmark.
We unify these strategies into a single objective by incorporating a regularization term based on the KL-divergence.
\begin{algorithm}
\small 
\caption{TAGS Sampling}
\label{alg:ours}
\begin{algorithmic}[1]
\STATE \textbf{Input:} Data Pool $D_P$, Sample Budget $N$, Target Distribution $Q$
\STATE \textbf{Output:} The Sampled Dataset $D_S$
\STATE Initialize Empty $D_S$;
\STATE Initialize Propagation Matrix $A$;
\WHILE{$|D_S| < N$}
    \STATE $G \gets \Phi'(A\sum_{k \in D_S} \mathbf{e}_k)^\top A$;
    \STATE $d_i \gets \mathop{\arg\max}_{d \in D_P} G \mathbf{e}_d - \lambda \mathcal{D}_{\text{KL}}(Q || P(D_S \cup \{d\}))$;
    \STATE $D_S \gets D_S \cup \{d_i\}$;
    \STATE $D_P \gets D_P \setminus \{d_i\}$;
\ENDWHILE
\RETURN $D_S$
\end{algorithmic}
\end{algorithm}
Defining the leaf-level distribution of the current subset as $P(D_S) \propto \sum_{d \in D_S} \mathbf{h}^{(\text{leaf})}_d$, 
we select the sample maximizing the joint score at iteration $k$:
\begin{equation}
\label{eq:final_obj}
\small
    d_k = \operatorname*{argmax}_{d \in D_P^k} ( \Delta \mathcal{I}(d) - \lambda \cdot \mathcal{D}_{\text{KL}}(Q \parallel P(D_S^k \cup \{d\})) ),
\end{equation}
where $\Delta \mathcal{I}(d)$ denotes the approximated gain from Eq.~\ref{eq:gain}.
Setting $\lambda=0$ reduces to general sampling, while $\lambda > 0$ enforces distribution alignment with the target $Q$.
The procedure is outlined in Algorithm~\ref{alg:ours}.
More details are shown in Appendix~\ref{details}.

\section{Experiments}

We validate the TAGS framework through extensive experiments across multiple benchmarks, following established settings~\cite{lu2023instag,chen2025mig}. Our evaluation follows a bottom-up design, addressed by three key research questions:
\begin{itemize}[leftmargin=*, nosep]
\item \textbf{RQ1:} Can the TAGS Tagger accurately capture instruction intent and knowledge semantics?
\item \textbf{RQ2:} Does the hierarchical Tag Tree exhibit a coherent and interpretable structure?
\item \textbf{RQ3:} Do the TAGS sampling strategies effectively improve SFT performance?
\end{itemize}

\subsection{Tasks and Datasets.}
\begin{figure}
    \centering
    \includegraphics[width=0.9\linewidth]{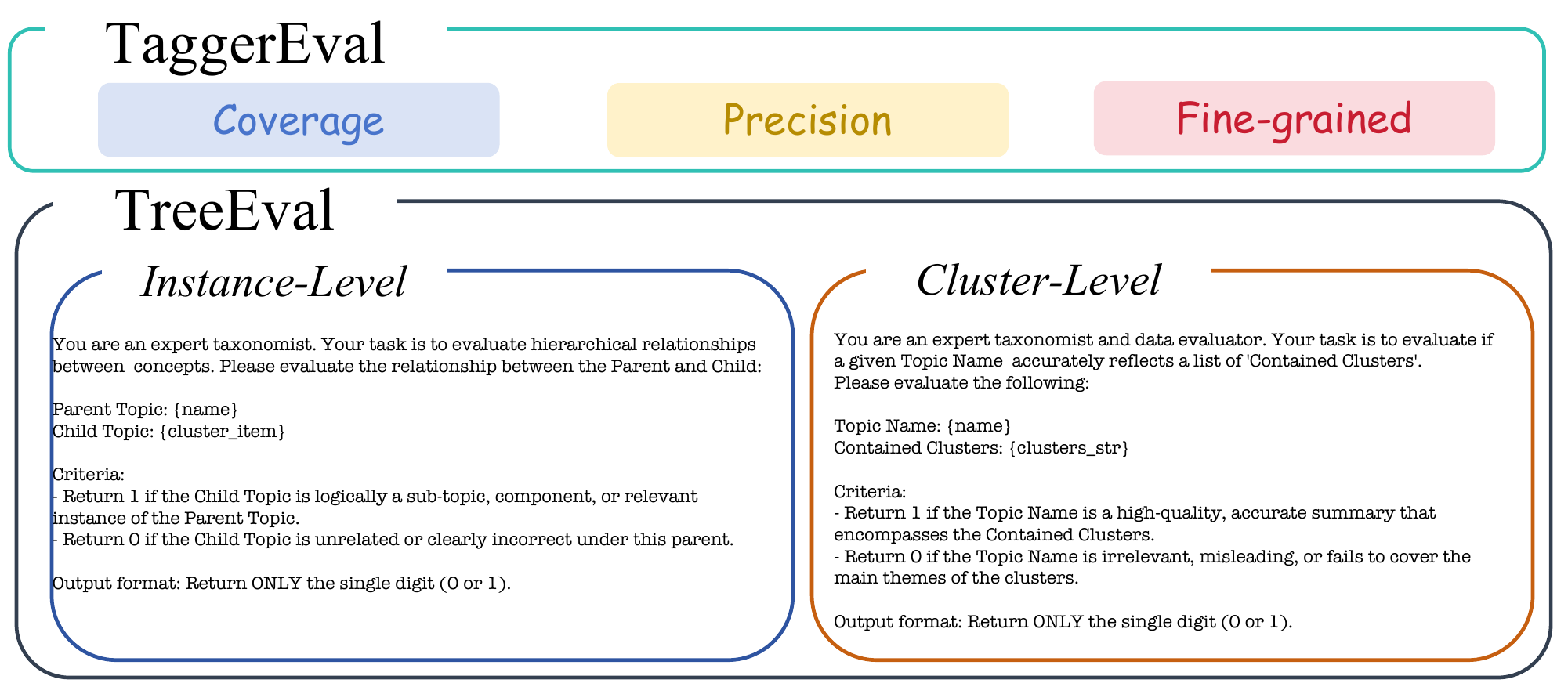}
    \caption{The TaggerEval and TreeEval Frameworks.}
    \label{fig:TagEval}
    \vspace{-1.5em} 
\end{figure}

\noindent
\textbf{Tagger Evaluation.}
To evaluate the proposed tagger, we introduce TaggerEval, a human-aligned evaluation framework (Figure~\ref{fig:TagEval}). We construct a benchmark of 100 high-quality samples spanning diverse domains, and use an powerful LLM\footnote{We utilize \texttt{GPT-4.1} as our judge model.} to assess the generated tags along three binary dimensions, following~\cite{lee2025checkeval}: 
\begin{itemize}[leftmargin=*, nosep]
\item \textbf{Coverage} evaluates whether each golden-truth concept is included in the generated tags.
\item \textbf{Precision} assesses whether each generated tag correctly reflects the input instruction.
\item \textbf{Fine-grained} measures whether each generated tag represents a fine-grained, atomic concept.
\end{itemize}
We report the average score on these metrics as the tagger performance. To ensure evaluation reliability, we also report Cohen’s kappa~\cite{sim2005kappa, 10.1145/3397271.3401334} between the LLM judge and human ratings.
\vspace{0.1em}

\noindent
\textbf{Tree Evaluation.}
To ensure the semantic coherence of our taxonomy, we introduce \textit{TreeEval}, as shown in Figure~\ref{fig:TagEval}. We construct a benchmark consisting of 200 randomly sampled tag-parent pairs and 50 parent clusters to assess structural quality from two perspectives:
\begin{itemize}[leftmargin=*, nosep]
\item \textbf{Instance-Level} verifies whether the tags are logical sub-topics of their assigned clusters.
\item \textbf{Cluster-Level} checks whether the cluster name captures the shared semantics of its children.
\end{itemize}
Similar to TaggerEval, we employ LLM to score these samples based on binary correctness.

\vspace{0.1em}
\noindent
\textbf{Data Sampling Evaluation.}
We conducted experiments on the Tulu3 dataset~\cite{lambert2024tulu} using Qwen3 models~\cite{yang2025qwen3} of various sizes. 
Model performance was comprehensively evaluated using knowledge-intensive and human-preference benchmarks. 
Evaluations were performed with OpenCompass~\cite{2023opencompass}, and average results are reported as normalized scores in percentage.

\begin{table*}[ht]
\centering
\scriptsize
\setlength{\tabcolsep}{5pt} 
\begin{tabular}{l | l l | c c c c c c | c | c c c | c | c}
\toprule
Base Model & Method & DataSize & ARC & BBH & GSM & HE & MMLU & IFEval & Avg\textsubscript{\textit{obj}} & AE & MT & Wild & Avg\textsubscript{\textit{sub}} & Avg \\
\midrule
\multirow{8}{*}{Qwen3-8B}
& Pool   & 939K & 88.47 & 69.71 & 88.86 & 77.44 & 75.27 & \textbf{69.87} & 78.27 & 9.69 & 7.34 & \underline{44.52} & 42.54 & 60.40 \\
\cmidrule{2-15}
& Random & 50K & 80.68 & 72.29 & 90.75 & \underline{82.93} & 78.60 & 64.14 & 78.23 & 9.44 & 7.51 & 41.74 & 42.09 & 60.16 \\
& IFD    & 50K & \textbf{90.85} & \textbf{77.65} & 91.36 & 81.71 & 78.79 & 57.12 & \underline{79.58} & 15.16 & 7.62 & 42.32 & 44.56 & \underline{62.07} \\
& \textit{\#InsTag} & 50K & 85.42 & 70.78 & 90.75 & 80.49 & 78.81 & 60.63 & 77.81 & 10.68 & \underline{7.67} & 42.16 & 43.18 & 60.50 \\
& DEITA  & 50K & 60.00 & 71.59 & \textbf{92.65} & \textbf{85.37} & 78.61 & 60.44 & 74.78 & 14.04 & 7.62 & 44.16 & 44.80 & 59.79 \\
& CaR    & 50K & 79.66 & 73.40 & 91.36 & 82.32 & \underline{78.91} & 56.19 & 76.97 & 13.29 & 7.49 & 42.52 & 43.57 & 60.27 \\
& QDIT   & 50K & 76.95 & \underline{77.40} & 90.67 & 81.71 & 78.82 & 58.23 & 77.30 & 14.04 & 7.41 & 40.90 & 43.01 & 60.16 \\
& MIG & 50K & 76.61 & 74.49 & \underline{91.51} & 79.88 & \textbf{79.04} & 65.80 & 77.89 & \underline{16.65} & 7.59 & 43.49 & \underline{45.35} & 61.46 \\
\cmidrule{2-15}
& Ours & 50K & \underline{86.78} & 74.22 & \underline{91.51} & 82.32 & 78.75 & \underline{66.91} & \textbf{80.08} & \textbf{20.12} & \textbf{7.85} & \textbf{44.76} & \textbf{47.79} & \textbf{63.93} \\
\midrule
\midrule
\multirow{7}{*}{Qwen3-4B}
& Random & 50K & 85.42 & 67.75 & \underline{89.23} & 78.05 & 75.29 & 61.18 & 76.15 & 9.07 & 7.25 & 38.16 & 39.91 & 58.03 \\
& IFD    & 50K & 91.53 & \underline{72.40} & 78.17 & 76.22 & 75.23 & 57.86 & 75.23 & \textbf{15.65} & 7.26 & 40.15 & 42.80  & 59.02 \\
& \textit{\#InsTag} & 50K & 84.07 & 66.88 & 86.96 & \underline{78.66} & 75.35 & 63.59 & 75.92 & 9.57 & \underline{7.38} & 38.26 & 40.54  & 58.23 \\
& DEITA  & 50K & 76.95 & 67.83 & 81.35 & \underline{78.66} & \textbf{75.56} & 63.22 & 73.93 & 8.20 & 7.26 & 38.14 & 39.64  & 56.78 \\
& CaR    & 50K & \underline{91.86} & 68.51 & 79.30 & 77.44 & 75.26 & 54.90 & 74.54 & 13.54 & 7.33 & 38.80 & 41.88 & 58.21 \\
& QDIT   & 50K & 91.19 & 72.90 & 85.06 & 76.83 & 75.16 & 58.78 & \underline{76.65} & 12.80 & 7.24 & 38.86 & 41.35 & 59.00 \\
& MIG & 50K & 91.53 & 72.10 & 70.43 & \textbf{81.10} & \underline{75.47} & \underline{65.80} & 76.07 & 14.04 & \textbf{7.64} & \underline{40.66} & \textbf{43.70}  & \underline{59.88} \\
\cmidrule{2-15}
& Ours & 50K & \textbf{92.54} & 70.78 & \textbf{89.69} & 76.83 & 75.33 & \textbf{66.73} & \textbf{78.65} & \underline{14.78} & 7.34 & \textbf{40.98} & \underline{43.05}  & \textbf{60.85} \\
\midrule
\midrule
\multirow{8}{*}{Qwen3-1.7B}
& Random & 50K & 78.64 & 51.68 & 77.26 & 65.85 & 64.33 & 42.14 & 63.32 & 4.84 & \underline{6.64} & 25.02 & 32.09  & 47.70 \\
& IFD    & 50K & 78.31 & 53.92 & 54.97 & 67.07 & 64.20 & 35.49 & 58.99 & 6.21 & 6.60 & 25.46 & 32.56  & 45.77 \\
& \textit{\#InsTag} & 50K & 78.31 & 50.88 & 77.26 & 66.46 & 64.27 & 43.44 & 63.44 & 5.09 & 6.51 & 26.17 & 32.12 & 47.78 \\
& DEITA  & 50K & 71.53 & 51.42 & 62.55 & 65.85 & 64.29 & 42.33 & 59.66 & 5.22 & 6.33 & 25.40 & 31.31  & 45.48 \\
& CaR    & 50K & 71.86 & 52.70 & 73.77 & 61.59 & \textbf{64.36} & 38.45 & 60.45 & 6.46 & 6.43 & 26.42 & 32.40 & 46.42 \\
& QDIT   & 50K & 76.95 & \textbf{54.62} & 58.07 & \underline{67.07} & 64.13 & 37.71 & 59.76 & 7.08 & 6.26 & 24.49 & 31.39 & 45.58 \\
& MIG & 50K & \underline{80.34} & 53.53 & \underline{78.77} & 66.46 & 64.00 & \underline{44.36} & \underline{64.58} & \underline{9.69} & 6.54 & \underline{27.92} & \underline{34.34}  & \underline{49.46} \\
\cmidrule{2-15}
& Ours & 50K & \textbf{82.03} & \underline{54.31} & \textbf{80.14} & \textbf{67.68} & 63.31 & \textbf{48.43} & \textbf{65.98} & \textbf{10.93} & \textbf{6.76} & \textbf{29.83} & \textbf{36.12}  & \textbf{51.05} \\
\bottomrule
\end{tabular}
\caption{Comparison of data selection methods on Tulu3 pool. Avg is mean of $Avg_{obj}$ and $Avg_{sub}$. TAGS achieves best performance across all base models.}
\label{tab:overall_results}
\vspace{-1.5em}
\end{table*}

\begin{itemize}[leftmargin=*, nosep]
\item \textbf{Knowledge-intensive Benchmarks:} We examine the model's factual knowledge, reasoning, coding, math, and instruction-following abilities across ARC~\cite{clark2018think}, BBH~\cite{suzgun2023challenging}, MMLU~\cite{hendrycks2020measuring}, HumanEval~\cite{chen2021evaluating}, GSM8k~\cite{cobbe2021training}, and IFEval~\cite{zhou2023instruction}.
\item \textbf{Human-preference Benchmarks:} We assess model's open-ended performance on AlpacaEvalv2~\cite{dubois2024length}, MTBench~\cite{zheng2023judging}, and WildBench~\cite{lin2024wildbench}.
\item \textbf{Baselines:} We compare our methods against strong data selection baseline including: random selection~\cite{xia2024rethinking}, IFD~\cite{li2024quantity}, \textit{\#InsTag}~\cite{lu2023instag}, DEITA~\cite{liumakes}, CaR~\cite{ge2024clustering}, QDIT~\cite{bukharin2024data} and MIG~\cite{chen2025mig}.
\end{itemize}

\begin{table}[ht]
\centering
\scriptsize
\setlength{\aboverulesep}{0pt}
\setlength{\belowrulesep}{0pt}
\renewcommand{\arraystretch}{1.2}
\setlength{\tabcolsep}{4pt}
\begin{tabular}{c|c c c c}
\toprule
\textbf{Method} & \textbf{Coverage} & \textbf{Precision} & \textbf{Fine-grained} & \textbf{Avg.} \\
\midrule
\textit{\#InsTagger} & 0.4527 & 0.2514 & 0.1286 & 0.2776 \\
Ours(w/o Act-Critic ) & 0.7387 & 0.7815 & 0.6328 & 0.7177 \\
Ours(w Act-Critic ) & \textbf{0.7623} & \textbf{0.817} & \textbf{0.7793} & \textbf{0.7862} \\
\hline
\rowcolor{gray!10}
\textit{kappa (vs human)} & 0.7069 & 0.7826 & 0.7218 & - \\
\bottomrule
\end{tabular}
\caption{Tag Performance Between Different Taggers.}
\label{tab:tag_results}
\vspace{-1.5em} 
\end{table}

\subsection{Main Results \& Analysis}
\subsubsection{Analysis of Tagger Performance}

\noindent\textit{\textbf{Q1: How does the proposed TAGS Tagger perform compare with baselines?}}\\
\noindent 
As shown in Table~\ref{tab:tag_results}, the TAGS Tagger consistently outperforms \textit{\#InsTagger} across all metrics, achieving a \textbf{+183.21\%} relative improvement in the average score and demonstrating substantial gains in coverage, precision, and granularity.

\vspace{0.5em}
\noindent\textit{\textbf{Q2: What is the impact of Act-Critic pipeline?}}\\
\noindent 
Ablation studies further confirm the necessity of the Act-Critic design. Compared to the direct prompting baseline (\textit{w/o Act-Critic}), employing the Act-Critic pipeline (\textit{w/ Act-Critic}) yields consistent gains across all dimensions (average gain of \textbf{+9.54\%}). This validates that the Act-Critic mechanism effectively enhance data synthesis performance, as validated in~\cite{madaan2023self}.

\subsubsection{Analysis of Hierarchical Clustering}
\textit{\textbf{Q3: Is the constructed Tag Tree reliable?}}

\begin{table}[ht]
\centering
\small
\setlength{\aboverulesep}{0pt}
\setlength{\belowrulesep}{0pt}
\renewcommand{\arraystretch}{1.3}
\setlength{\tabcolsep}{4pt}
\begin{tabular}{c|c c}
\toprule
\textbf{Method} & \textbf{Cluster-Level} & \textbf{Instance-Level} \\
\midrule
Hierarchical Clustering & 98.00\% & 96.00\%  \\
\bottomrule
\end{tabular}
\caption{Hierarchical Clustering Performance.}
\label{tab:tree_results}
\vspace{-1.5em} 
\end{table}

\noindent 
Results in Table~\ref{tab:tree_results} demonstrate the high reliability of our taxonomy. The constructed hierarchy attains strong validation scores, with 98.00\% accuracy at Cluster level and 96.00\% at Instance level. These results indicate that the parent–child relations are both logically sound and semantically coherent.

\subsubsection{Analysis of Data Sampling}

\begin{figure*}
\centering
\includegraphics[width=0.8\linewidth]{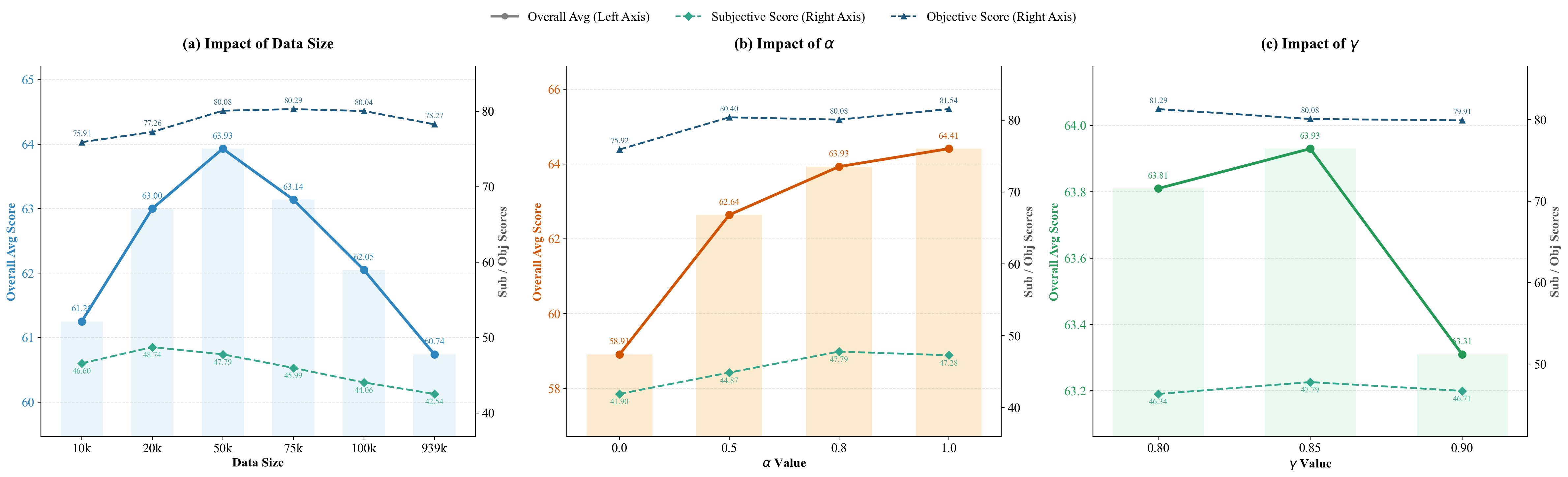}
\caption{Ablation studies on training data size, hyperparameter $\alpha$ of complicated score $s$, and $\gamma$ value of $\Phi(x)$.
}
\label{fig:ablation}
\vspace{-1.0em}
\end{figure*}

\noindent\textit{\textbf{Q4: How does the  General Sampling perform regarding overall effectiveness and data efficiency?}}

\noindent As illustrated in Table~\ref{tab:overall_results}, TAGS consistently outperforms all baselines across objective and subjective metrics. 
Specifically, on the Qwen3-8B backbone, TAGS achieves the highest average score, surpassing the strongest baseline by \textbf{+2.99\%} and the Random selection baseline by \textbf{+6.27\%}.
Notably, while maintaining competitive capability on knowledge benchmarks (\textbf{+0.63\%}), TAGS delivers a substantial breakthrough in human-preference alignment (\textbf{+5.38\%}) compared to the strongest baseline. 
This suggests that the general-purpose sampled, diverse data significantly improves the model’s performance on open-ended instructions.
Crucially, TAGS demonstrates a profound \textit{``less is more''} phenomenon:
by utilizing only \textbf{5\%} of the data pool, it outperforms the model trained on the full dataset by \textbf{+5.84\%}.
\begin{table}[ht]
\centering
\scriptsize
\setlength{\aboverulesep}{0pt}
\setlength{\belowrulesep}{0pt}
\renewcommand{\arraystretch}{1.2}
\setlength{\tabcolsep}{5pt}
\begin{tabular}{c | c | c c c}
\toprule
\textbf{Method} & \textbf{Data Size} & \textbf{ARC} & \textbf{GSM8K} & \textbf{GPQA-Diamond} \\
\midrule
TAGS        & 10K & 91.19 & 85.67 & 30.30 \\
TAGS-\textit{Align}  & 10K & 93.90 & 89.69 & 31.83 \\
\hline
\rowcolor{gray!10}
Improvement  & - & \textbf{+2.97\%} & \textbf{+4.69\%} &  \textbf{+5.06\%} \\
\bottomrule
\end{tabular}
\caption{Performance of aligned sampling strategy.}
\label{tab:gpqa}
\vspace{-1.5em}
\end{table}
This confirms that the tree-aware sampling effectively identifies high-quality, diversity samples, eliminating redundancy and noise.

\noindent\textit{\textbf{Q5: Is the proposed strategy scalable across different model sizes?}}

\noindent TAGS demonstrates strong scalability, consistently ranking first across 1.7B, 4B, and 8B models, as shown in Table~\ref{tab:overall_results}. Notably, it yields the largest gains on the 1.7B model, improving the average score by \textbf{+3.21\%} over the best baseline, indicating its effectiveness independent of model scale.

\vspace{0.2em}
\noindent\textit{\textbf{Q6: Is the Aligned Sampling strategy effective across diverse benchmarks?}}

\noindent 
To verify adaptability, 
we map specific benchmarks (ARC~\cite{clark2018think}, GSM8K~\cite{cobbe2021training}, and GPQA-Diamond~\cite{rein2024gpqa}) to Tag Tree and perform aligned sampling (set $\lambda=5$) with a budget of 10k samples.
As shown in Table~\ref{tab:gpqa}, this strategy yields comprehensive improvements, boosting average performance by \textbf{+4.24\%}.
These results indicate that TAGS effectively aligns sampled data with diverse, target domains, enabling fine-grained steering of model capabilities toward specific domain constraints or user requirements.

\subsection{Ablation Study}
\noindent\textit{\textbf{Q7: What are the contributions of the fine-grained tagger and the tree structure?}}

\noindent To decouple the contributions of our key components, we compare three configurations as detailed in Table~\ref{tab:ablation}:
(1) \textbf{MIG}, which serves as the baseline using coarse-grained tags without tag tree;
(2) \textbf{TAGS (w/o Tree)}, which employs our fine-grained tagger but relies on flat label graph;
and (3) \textbf{TAGS}, which integrates both the fine-grained tagger and the hierarchical tag tree.
\begin{table}[ht]
\centering
\scriptsize
\setlength{\aboverulesep}{0pt}
\setlength{\belowrulesep}{0pt}
\renewcommand{\arraystretch}{1.2}
\setlength{\tabcolsep}{5pt}
\begin{tabular}{c | c | c c | c}
\toprule
\textbf{Method} & \textbf{DataSize} & \textbf{Avg\textsubscript{\textit{obj}}} & \textbf{Avg\textsubscript{\textit{sub}}} & \textbf{Avg} \\
\midrule
MIG & 50k & 77.89  & 45.35 & 61.46 \\
TAGS(w/o Tree) & 50k & \underline{80.05}                & \underline{47.00} & \underline{63.52} \\
\hline
\rowcolor{gray!10}
TAGS & 50k & \textbf{80.08} & \textbf{47.79} & \textbf{63.93} \\
\bottomrule
\end{tabular}
\caption{Ablation study of TAGS.}
\vspace{-1.5em}
\label{tab:ablation}
\end{table}
The results reveal two key insights. First, the transition from coarse to fine-grained tagging (\textbf{MIG} vs. \textbf{TAGS w/o Tree}) yields a substantial performance leap of \textbf{+3.35\%}, underscoring the critical importance of atomic data representation.
Second, incorporating the hierarchical structure (\textbf{TAGS}) provides further optimization, particularly in subjective alignment, where it boosts performance by \textbf{+1.68\%}, raising the final average score by \textbf{+0.64\%}. 
These findings suggest that fine-grained tags enable precise data assessment, while the tag tree ensures diversity and balance.
\vspace{0.2em}
\noindent\textit{\textbf{Q8: How do key hyperparameters and data scale affect the effectiveness of TAGS?}}

\noindent Figure~\ref{fig:ablation} shows that TAGS performance peaks at 50K samples, indicating that selection is more critical than sheer scaling. While setting $\alpha=1.0$ improves average scores, incorporating complexity ($\alpha=0.8$) yields the best results on challenging reasoning tasks (BBH, MMLU) and open-ended benchmarks (MT-Bench, WildBench), suggesting that a 20\% complexity weight helps select harder, more informative samples. 
Additionally, setting \textbf{$\gamma=0.85$} balances noise filtering with sample retention. Detailed results are shown in the Table~\ref{tab:alpha_comparison}.

\section{Conclusion}
In this paper, we present TAGS, a framework that integrates fine-grained tags into a hierarchical tree for controllable data selection. By organizing atomic tags into a global taxonomy, TAGS enables sampling that balances data quality and diversity. Extensive experiments show that TAGS consistently outperforms strong baselines across diverse benchmarks, achieving superior performance with only 5\% of the original data, demonstrating its efficacy for efficient instruction tuning.

\section*{Limitations}
\begin{itemize}
\item[1.] \textbf{Tagger Data Scale:} To balance performance with proprietary API costs, our Tagger was fine-tuned on a compact dataset of 53K samples. While this effectively validates data efficiency, it precludes a comprehensive analysis of the Tagger's scaling behavior on larger synthetic datasets.
\item[2.] \textbf{Model \& Task Scope:} Due to computational constraints, our evaluation focuses on the \texttt{Qwen3-1.7B, 4B, 8B} series for SFT tasks. Verification on larger-scale models (e.g., \texttt{Qwen3-30B}) and other paradigms like Reinforcement Learning remains for future exploration.
\end{itemize}
\bibliography{custom}

\appendix

\section{More TAGS Implementation Details}
\label{sec:appendix}
\subsection{Baseline Settings}

We adhere to the comparison settings established by MIG~\cite{chen2025mig} and utilize 50k sampled instances from each MIG-released baseline method\footnote{\url{https://huggingface.co/collections/xsample/mig-datasets}} to conduct a consistent and fair SFT evaluation. This setup ensures that all methods are assessed under uniform data conditions, enabling a direct comparison of their intrinsic capabilities.

\subsection{Implementation Details}
\label{details}
\subsubsection{Tagger Training.} 
We first collect 53k Supervised Fine-Tuning samples synthesized by \texttt{GPT-o4-mini} using an Actor–Critic pipeline to train the TAGS Tagger, and then further fine-tune it on \texttt{Qwen3-8B-Instruct}\footnote{\url{https://huggingface.co/Qwen/Qwen3-8B}}
.
The distribution of the training data is summarized in Table~\ref{tab:data_domain}.
For non-complex queries, we apply uniform sampling over the entire pool.
For complex queries, we perform uniform sampling from several inherently high-difficulty datasets spanning the Math, Code, and STEM domains:
\begin{itemize}[leftmargin=*, nosep]
\item \textbf{Math}: AIME and ArenaHard-Math.
\item \textbf{Code}: ArenaHard-Coding.
\item \textbf{STEM}: SuperGPQA.
\end{itemize}

\subsubsection{Tree Constructing.} 
We adopt the bottom‑up hierarchical clustering algorithm described in Clio~\cite{tamkin2024clio}. Starting from 7M high‑quality open‑source samples~\cite{OpenHermes,xu2024magpie,numina_math_datasets,xu-etal-2025-kodcode,li2025infinity}, we first apply the TAGS Tagger to extract fine‑grained tags. To ensure computational efficiency at scale, we use all‑mpnet‑base‑v2\footnote{\url{https://huggingface.co/sentence-transformers/all-mpnet-base-v2}} as the encoder to perform an initial coarse‑grained clustering, reducing the data into 10,000 preliminary clusters.
We then employ \texttt{Qwen3‑30B-A3B-2507}\footnote{\url{https://huggingface.co/Qwen/Qwen3-30B-A3B-Thinking-2507}} to summarize, deduplicate, reassign, and rename cluster labels, producing cleaner and more semantically coherent tag groups. After this refinement step, we further apply Qwen3‑Embedding‑0.6B\footnote{\url{https://huggingface.co/Qwen/Qwen3-Embedding-0.6B}} to conduct subsequent clustering, which provides a better balance between clustering quality and computational cost.
This multi‑stage pipeline ultimately yields a 10‑level hierarchical tree, offering a structured and interpretable organization of the underlying data space.

\subsubsection{TAGS Sampling \& Training.} 
We sample 50k training instances from the full 939k Tulu3 dataset~\cite{lambert2024tulu} and use 8 GPUs to accelerate both the sampling process and the subsequent SFT training. 
In the initial stage of sampling, we initialize $G$ as a zero vector.
For training stage, we set the batch size to 128, learning rate to 5e‑6, warmup ratio to 0.03, and the maximum length to 8192. 
The implementation is conducted using the Swift framework~\cite{zhao2024swiftascalablelightweightinfrastructure}.
As shown in Table~\ref{tab:time_cost}, General Sampling is extremely efficient, occupying less than 0.2\% of the full training time while selecting a high-quality 50k subset. Even for Aligned Sampling, which involves complex KL-divergence calculations, the computational overhead remains minimal, ensuring practical usability.
Additionally, our sampling strategy compresses the total training time to 5\%, yet achieves superior performance, fully demonstrating its efficiency and practical value.

\begin{table}[ht]
    \centering
    \begin{tabular}{c|c}
    \toprule
         \textbf{Stage} & \textbf{Time}(min) \\
         \midrule
         General Sampling &  3.13\\
         Aligned Sampling & 35.51 \\
         SFT(50k Data) & 90.85 \\
         SFT(939k Data) & 1810.53 \\
    \bottomrule
    \end{tabular}
    \caption{Computational Overhead of Sampling and Training Phases.}
    \label{tab:time_cost}
\end{table}

\section{Detailed Experimental Results}
We present detailed experimental results across varying data scales (Table~\ref{tab:data_size}), hyperparameter $\alpha$ (Table~\ref{tab:alpha_comparison}), hyperparameter $\gamma$ (Table~\ref{tab:k_comparison}), and ablation study of fine-grained Tagger and Tag Tree (Table~\ref{tab:ablation}).

\begin{table*}[t]
    \centering
    \scriptsize
    \begin{tabular}{l  c  c}
        \toprule
        \textbf{Domain} & \textbf{Quantity} & \textbf{Source} \\
        \midrule
        General & 10K & Magpie-Pro~\cite{xu2024magpie}, Infinity-Instruct~\cite{li2025infinity}, OpenHermes~\cite{OpenHermes}, \dots \\
        \midrule
        Math    & 10K & NuminaMath-1.5~\cite{numina_math_datasets}, DeepMath-103K~\cite{he2025deepmath}, Aqua-Rat~\cite{ling2017program}, \dots \\
        \midrule
        Code    & 10K & TACO~\cite{li2023taco}, KODCode~\cite{xu-etal-2025-kodcode}, LeetCodeDataset~\cite{xia2025leetcodedataset}, \dots \\
        \midrule
        STEM    & 10K & Natural-Reasoning~\cite{yuan2025naturalreasoningreasoningwild28m}, MegaScience~\cite{xu2024magpie},  SciRiFF~\cite{wadden-etal-2025-sciriff}, \dots \\
        \midrule
        Chat    & 10K & lmsys-Chat~\cite{zheng2023lmsyschat1m} \\
        \midrule
        Complex & 3k & AIME~\cite{aime_1983_2024}, ArenaHard(Math, Coding)~\cite{li2024crowdsourced}, SuperGPQA~\cite{pteam2025supergpqascalingllmevaluation}, \dots \\
        \bottomrule
    \end{tabular}
    \caption{Domain distribution and source datasets used in tagger training.}
    \label{tab:data_domain}
\end{table*}

\begin{table*}[ht]
\setlength{\aboverulesep}{0pt}
\setlength{\belowrulesep}{0pt}
\renewcommand{\arraystretch}{1.2}
\setlength{\tabcolsep}{5pt}
\scriptsize
\centering
\begin{tabular}{l | c | c c c c c c | c | c c c | c | c}
\toprule
Method & DataSize & ARC & BBH & GSM & HE & MMLU & IFEval & Avg\textsubscript{\textit{obj}} & AE & MT & Wild & Avg\textsubscript{\textit{sub}} & Avg \\
\midrule
\multirow{5}{*}{TAGS}
& 10k  & \textbf{91.19} & 70.15 & 85.67 & 69.51 & 78.49 & 60.44 & 75.91 & 18.63 & \underline{7.74} & 43.77 & 46.60 & 61.25 \\
& 20k  & 85.42 & 70.16 & 86.88 & 79.27 & 76.74 & 65.06 & 77.26 & \textbf{22.11} & \textbf{7.85} & \textbf{45.62} & \textbf{48.74} & 63.00 \\
& 50k  & 86.78 & \textbf{74.22} & \underline{91.51} & \textbf{82.32} & \underline{78.75} & 66.91 & \underline{80.08} & \underline{20.12} & \textbf{7.85} & \underline{44.76} & \underline{47.79} & \textbf{63.93} \\
& 75k  & 86.78 & \underline{74.00} & \underline{91.51} & \textbf{82.32} & 78.74 & 68.39 & \textbf{80.29} & 17.02  & 7.68 & 44.16 & 45.99 & \underline{63.14} \\
& 100k & 84.07 & 73.27 & \textbf{91.66} & \underline{81.10} & \textbf{79.01} & \textbf{71.16} & 80.04 & 15.78  & 7.38 & 42.58 & 44.06 & 62.05 \\
\hline
\rowcolor{gray!20}
ALL & 939k & \underline{88.47} & 69.71 & 88.86 & 77.44 & 75.27 & \underline{69.87} & 78.27 & 9.69  & 7.34 & 44.52 & 42.54 & 60.74 \\
\bottomrule
\end{tabular}
\caption{Performance comparison of different data size on objective and subjective benchmarks.}
\label{tab:data_size}
\end{table*}

\begin{table*}[ht]
\setlength{\aboverulesep}{0pt}
\setlength{\belowrulesep}{0pt}
\renewcommand{\arraystretch}{1.2}
\setlength{\tabcolsep}{5pt}
\scriptsize
\centering
\begin{tabular}{l | c | c c c c c c | c | c c c | c | c}
\toprule
Method & DataSize & ARC & BBH & GSM & HE & MMLU & IFEval & Avg\textsubscript{\textit{obj}} & AE & MT & Wild &Avg\textsubscript{\textit{sub}} & Avg \\
\midrule
TAGS ($\alpha$=0.0) & 50K
    & 80.34 & 73.82 & 78.24 & 84.15 & 78.89 & 60.07 & 75.92 & 10.93 & 7.28 & 41.96 & 41.90 & 58.91 \\
TAGS ($\alpha$=0.5) & 50K
    & 85.08 & 73.59 & 91.13 & \textbf{84.76} & \textbf{78.89} & \underline{68.95} & \underline{80.40} & 16.02 & 7.53 & 43.29 & 44.87 & 62.64 \\
TAGS ($\alpha$=0.8) & 50K
    & \underline{86.78} & \textbf{74.22} & \underline{91.51} & 82.32 & \underline{78.75} & 66.91 & 80.08 & \underline{20.12} & \textbf{7.85} & \textbf{44.76} & \textbf{47.79} & \underline{63.93} \\ 
TAGS ($\alpha$=1.0) & 50K
    & \textbf{87.80} & \underline{74.12} & \textbf{92.19} & \underline{84.15} & 78.71 & \textbf{72.27} & \textbf{81.54} & \textbf{20.75} & \underline{7.67} & \underline{44.40} & \underline{47.28} & \textbf{64.41}\\
\bottomrule
\end{tabular}
\caption{Performance comparison of different hyperparameter $\alpha$ on objective and subjective benchmarks.}
\label{tab:alpha_comparison}
\end{table*}

\begin{table*}[ht]
\setlength{\aboverulesep}{0pt}
\setlength{\belowrulesep}{0pt}
\renewcommand{\arraystretch}{1.2}
\setlength{\tabcolsep}{5pt}
\scriptsize
\centering
\begin{tabular}{l | c | c c c c c c | c | c c c | c | c}
\toprule
Method & DataSize & ARC & BBH & GSM & HE & MMLU & IFEval & Avg\textsubscript{\textit{obj}} & AE & MT & Wild & Avg\textsubscript{\textit{sub}} & Avg \\
\midrule
TAGS ($\gamma = 0.80$) & 50K & \textbf{89.83}	& 73.86	 & \textbf{92.42} &	\textbf{82.93}	& \textbf{79.02}	& \textbf{69.69}	 & \textbf{81.29} &	19.13 &	7.55 &	44.38 &	46.34 &	\underline{63.81}
 \\
TAGS ($\gamma = 0.85$) & 50K & \underline{86.78} & \underline{74.22} & \underline{91.51} & \underline{82.32} & \underline{78.75} & 66.91 & \underline{80.08} & \textbf{20.12} & \textbf{7.85} & \underline{44.76} & \textbf{47.79} & \textbf{63.93} \\
TAGS ($\gamma = 0.90$) & 50K & 86.44 &	\textbf{75.02} &	91.05 &	78.66	& 78.79 &	\underline{69.50} &	79.91 &	\underline{19.38} &	\underline{7.59} &	\textbf{44.86} &	46.71 &	63.31
 \\
\bottomrule
\end{tabular}

\caption{Performance comparison of different hyperparameter $\gamma$ on objective and subjective benchmarks.}
\label{tab:k_comparison}
\end{table*}

\begin{table*}[ht]
\setlength{\aboverulesep}{0pt}
\setlength{\belowrulesep}{0pt}
\renewcommand{\arraystretch}{1.2}
\setlength{\tabcolsep}{5pt}
\scriptsize
\centering
\begin{tabular}{l | c | c c c c c c | c | c c c | c | c}
\toprule
Method & DataSize & ARC & BBH & GSM & HE & MMLU & IFEval & Avg\textsubscript{\textit{obj}} & AE & MT & Wild & Avg\textsubscript{\textit{sub}} & Avg \\
\midrule
MIG & 50K 
    & 76.61 & \textbf{74.49} & 91.51 & 79.88 & \textbf{79.04} & 65.80
    & 77.89       & 16.65 & \underline{7.59} & 43.49 & 45.35 & 61.46 \\
TAGS(w/o Tree) & 50K 
    & \underline{85.08} & 74.05 & \textbf{91.58} & \textbf{83.54} & \underline{78.98} & \textbf{67.10} 
    & \underline{80.05}                & \textbf{20.25} & 7.58 & \textbf{44.95} & \underline{47.00} & \underline{63.52} \\
TAGS & 50K 
    & \textbf{86.78} & \textbf{74.22} & \underline{91.51} & \underline{82.32} & 78.75 & \underline{66.91} & \textbf{80.08} & \underline{20.12} & \textbf{7.85} & \underline{44.76} & \textbf{47.79} & \textbf{63.93} \\
\bottomrule
\end{tabular}
\caption{Detailed performance of ablation study on objective and subjective benchmarks.}
\label{tab:method_comparison}
\end{table*}

\section{Prompt Template}
\subsection{Tagging Data Synthesis}
We instruct an LLM to act as GPT-Tagger for initial tagging (Table~\ref{tab:gpt_tagger}), then act as a GPT-Checker for tag refinement (Table~\ref{tab:gpt_checker}) to synthesize high-quality tagger training data. 
\begin{table*}
\begin{tcolorbox}[colback=gray!10,
                  colframe=black,
                  width=1\textwidth,
                  arc=1 mm, 
                  boxrule=0.5pt,
                  title=Instruction template for GPT-Tagger,
                  fontupper=\scriptsize
                 ]
You are a general knowledge extraction expert. Given a problem and its corresponding solution (which can be code, natural language, or other forms), your task is to identify all the fine-grained and atomic-level knowledge components involved.\\

\#\# Improvement Guidance\\
- Previous Tag: \{old\_tags\}\\
- Improvement Hint: \{improvement\_reason\}
  - If "None": This is a cold start, generate tags normally.\\
  - If not "None": This is feedback from the checker, refine your output accordingly. \\
  - For tags without explicitly identified issues, please keep them as they are. Only adjust the parts where problems are indicated.\\

  You must consider multiple dimensions, including but not limited to:\\
  * **Problem Domain** (e.g., "Number Theory", "Greedy Algorithms", "Ancient Rome History"),\\
  * **Problem Type** (e.g., "Calculation", "Proof", "Code Implementation", "Conceptual Explanation", "Dialogue Response"),\\
  * **Knowledge / Skill Points** (e.g., "Greatest Common Divisor", "Prime Factorization", "Binary Search", "Close Reading"),\\
  * **User Intent** (e.g., "Learning a concept", "Practical application", "Exam preparation"),\\

  Each knowledge point must be:\\
  * **Atomic**: the smallest meaningful concept or technique involved.\\
  * **Precise**: no vague categories like "Math" or "Data Structures".\\
  * **Unabbreviated**: e.g., use "Dynamic Programming" instead of "DP".\\
  * **Non-overlapping**: do not include multiple tags that describe the same idea with slight variation; only one representative tag should be included.\\
  * **Core-only**: include only the most essential knowledge points necessary to understand or implement the solution. Do not include secondary or peripheral techniques unless they are critical.\\
  * **Specifically Naming**: Your tag should be specific, include the core concept and entities into the tag.\\
  * **Focus** on the user’s primary intent while ignoring irrelevant context.\\
  * If the problem is in a multi-turn dialogue format, use the previous conversation history as contextual background, but focus primarily on the current query.\\
  * **Maximally concise**: output no more than *5* knowledge points.\\

  \#\#\# Output Format\\
  Return your results as a list of dictionaries, where each dictionary includes\\
  * `'tag'`: the name of the knowledge point\\
  * `'explanation'`: a brief and accurate explanation of its role in the problem and solution\\

  **Example**\\
  Problem: Find the GCD of 84 and 60\\
  Reference Answer: Let's find the **GCD (Greatest Common Divisor)** of $84$ and $60$. \#\#\# Step 1: Prime factorization \\
  * $84 = 2 \times 2 \times 3 \times 7 = 2^2 \times 3^1 \times 7^1$\\* $60 = 2 \times 2 \times 3 \times 5 = 2^2 \times 3^1 \times 5^1$\\
  \#\#\# Step 2: Take the lowest powers of common primes\\
  Common primes: $2$ and $3$\\
  * $2^{{\min(2,2)}} = 2^2 = 4$\\* $3^{{\min(1,1)}} = 3^1 = 3$\\
  Now multiply:\
  $$\\4 \times 3 = 12\\$$
  **GCD(84, 60) = 12**\\
  Output:
\begin{verbatim}
[{"tag": "Greatest Common Divisor", "explanation": "The problem requires computing the largest integer that divides 84 and 60."},
  {"tag": "Prime Factorization", "explanation": "One method of finding the GCD involves decomposing numbers into prime factors."},
  {"tag": "Euclidean Algorithm", "explanation": "An efficient iterative method to compute the GCD without full factorization."}]
\end{verbatim}

  Problem: Where is the Peking University?\\
  Reference Answer: No. 5 Yiheyuan Road, Haidian District, Beijing, China.\\
  Output:
\begin{verbatim}
[{"tag": "Peking University Location", "explanation": "Identifying the geographical location of Peking University, 
including its city and district."},
{"tag": "Beijing Geography", "explanation": "Knowledge of the districts of Beijing to contextualize the university's location."},
{"tag": "Higher Education Institutions in China", "explanation": "Understanding the placement and distribution of 
major universities in China for reference."}]
  \end{verbatim}

Problem:Which Sun sign, Rising sign, and Moon sign does a person born on September 17 (Gregorian calendar) belong to? Why are there differences among them?\\
Reference Answer:Sun sign: Virgo (because September 17 falls within the Virgo date range). Rising sign (Ascendant): Requires the **exact time and place of birth** to determine. Moon sign: Also requires the **exact time and place of birth** to calculate. The Sun sign depends only on the birth date, while the Rising sign and Moon sign vary with the **specific birth time and location**, as they change much more rapidly.\\
  Output:
  {\scriptsize
\begin{verbatim}
[{"tag": "Sun Sign Determination", "explanation": "Identifying the sun sign based on the Gregorian date of birth"},
{"tag": "Ascendant Sign Calculation", "explanation": "Determining the rising sign based on precise birth time and location"},
{"tag": "Moon Sign Calculation", "explanation": "Determining the moon sign based on the moon's position at the time of birth"},
{"tag": "Astrological Principles", "explanation": "Understanding why sun sign, moon sign, and ascendant can 
differ due to celestial mechanics and astrological rules"}]
  \end{verbatim}
  }
      **Now analyze the following input and return the list of atomic knowledge points. You should only output the final json. **\\

\#\#\# Problem:\\

\{problem\}\\

\#\#\# Reference Answer:\\

\{reference\_answer\}\\

\#\#\# Output:
'''
\end{tcolorbox}
\caption{Instruction template for GPT-Tagger Eval.}
\label{tab:gpt_tagger}
\end{table*}

\begin{table*}
\begin{tcolorbox}[colback=gray!10, colframe=black, width=1\textwidth, arc=1 mm, boxrule=0.5pt, title=Instruction template for GPT-Checker, fontupper=\small]
You are a tag validation expert. \\
You are given a problem description, reference answer and the extracted tag results. \\
Your task is to **evaluate the tag results** based on the following criteria:\\

\#\# Validation Criteria\\
1. **Output Format**: \\
   - Must be a JSON object.\\
   - "tags" must be a list of dictionaries, each with fields "explanation" and "tag".\\

2. **Content Reasonableness**:\\
   - Tags must match the problem description.\\
   - Tags must be meaningful and reasonably specific.\\

3. **Accuracy and Granularity**:\\
   - No vague categories (e.g. "Math" or "Data Structures").\\
   - Tags should not be overly broad (e.g. "Probability") nor overly fine-grained (e.g. "apply addition in Bayes formula").\\
   - Aim for an appropriate middle-level granularity (e.g. "Discrete Event Probability", "Binary Search Tree Traversal", "Video file handling in Python").\\
   - Tags must be correct, contains the core constraints and relevant to the problem.\\

4. **Coverage**:\\
   - Tags should cover the **essential core knowledge points** required to understand or solve the problem.\\
   - Maximum 5 tags; unnecessary extra tags should not appear.\\

\#\# Your Task\\
- If the output fully satisfies the criteria, respond with:\\
  {{"check":"Yes"}}\\

- If the output fails in any aspect, respond with:\\
  {{"check":"No", "Reason": "<concise reason for failure, with direct guidance to refine the tagging>"}}\\

---

\#\#\# Problem:\\
\{problem\_description\}\\

\#\#\# Reference Answer:\\

\{reference\_answer\}

\#\#\# Extracted Tags:\\
\{tag\_output\}
\end{tcolorbox}
\caption{Instruction template for GPT-checker.}
\label{tab:gpt_checker}
\end{table*}

\subsection{TaggerEval}
We prompt an LLM as a judge to evaluate the generated tags from coverage (Table~\ref{tab:coverage_eval}), precision (Table~\ref{tab:precision_eval}), fine-grained (Table~\ref{tab:finegrained_eval}) dimensions.

\begin{table*}
\begin{tcolorbox}[colback=gray!10,
                  colframe=black,
                  width=1\textwidth,
                  arc=1 mm, 
                  boxrule=0.5pt,
                  title=Instruction template for Coverager Eval,
                  fontupper=\small
                 ]
You are given two tag lists:

    1. **Raw Tags**: These are the ground-truth tags that represent concepts or labels in the data.\\
    2. **Generated Tags**: These are tags automatically generated by a system.\\\\
    Your task is to check whether each raw tag is **covered** by the generated tags.
    Please determine whether each raw tag was covered by the generated tag. \\\\
    Coverage criteria include:\\
1. Synonymous or closely related expressions
If the generated tag expresses the same or very similar meaning as the raw tag (including synonyms or paraphrases).\\
2. Containment relationship
If the generated tag directly contains the core concept of the raw tag, it counts as coverage.For example, "subarray counting" covers "array".\\
3. Hypernym-hyponym coverage
If the generated tag is a more specific concept or includes the core meaning of the raw tag, it is considered coverage. For example, "sliding window algorithm" covers "array".\\
4. Consistent domain terminology
Different expressions of the same technical term count as mutual coverage. For example, "greatest common divisor" and "GCD" cover each other.\\
5. Composite tag inclusion
If the generated tag is a composite concept and its core includes the raw tag’s core term with precise semantics, it counts as coverage. For example, "quick sort algorithm" covers "sorting".\\

Excluded cases (do not count as coverage):\\
Overly broad or generic tags are not considered coverage, even if they belong to the same general domain.\\
The generated tag must express a specific concept that directly includes or aligns with the raw tag’s meaning.\\

\#\#\# Output format:\\
Return a dictionary, where each key is a raw tag, and the value is `1` if it is covered by any generated tag, otherwise `0`.\\\\
\#\#\# Example:\\
Raw Tags: ["Named Entity Recognition", "POS Tagging", "DFS"]\\
Generated Tags: ["Part-of-Speech Tagging", "NER"]\\
Output:
\{\{
  "Named Entity Recognition": 1,
  "POS Tagging": 1,
  "DFS": 0
\}\}\\\\
You should output the dict in str format only, no analysis. 
Please carefully consider the intrinsic relationship between the generated tag and the raw tag, not just based on surface-level meaning. You may also make reasonable inferences — for example, character-level operations can be considered part of the string category.\\\\
Now evaluate:\\
Raw Tags: \{raw\_tags\}\\
Generated Tags: \{generate\_tags\}\\
\end{tcolorbox}
\caption{Instruction template for Coverager Eval.}
\label{tab:coverage_eval}
\end{table*}

\begin{table*}
\begin{tcolorbox}[colback=gray!10,
                  colframe=black,
                  width=1\textwidth,
                  arc=1 mm, 
                  boxrule=0.5pt,
                  title=Instruction template for Fine-grained Eval,
                  fontupper=\small
                 ]
You are given two tag lists:\\
    1. **Raw Tags**: These are the ground-truth tags that represent concepts or labels in the data.\\
    2. **Generated Tags**: These are tags automatically generated by a system.\\\\
    Your task is to check whether each generated tag is **accurate**, meaning it correctly corresponds to at least one raw tag. \\\\
    Accuracy criteria include:\\
    1. Synonymous or closely related expressions\\
    If the generated tag expresses the same or very similar meaning as any raw tag (including synonyms or paraphrases).\\
    2. Containment relationship\\
    If the generated tag directly contains the core concept of any raw tag, it counts as accurate. For example, "subarray counting" is accurate if the raw tags contain "array".\\
    3. Hypernym-hyponym relationship\\
    If the generated tag is a more specific concept or includes the core meaning of a raw tag, it is considered accurate.\\
    4. Consistent domain terminology\\
    Different expressions of the same technical term count as mutual accuracy. For example, "greatest common divisor" and "GCD" are accurate matches.\\
    5. Composite tag inclusion\\
    If the generated tag is a composite concept and its core includes the raw tag’s core term with precise semantics, it counts as accurate.\\\\
    Excluded cases (do not count as accurate):\\
    - Overly broad or generic generated tags that do not specifically match any raw tag.\\
    - Tags unrelated to any raw tag in meaning.\\

    \#\#\# Output format:\\
    Return a dictionary, where each key is a generated tag, and the value is `1` if it is accurate, otherwise `0`.\\

    \#\#\# Example:\\
    Raw Tags: ["Named Entity Recognition", "POS Tagging", "DFS"]\\
    Generated Tags: ["Part-of-Speech Tagging", "NER", "Graph Algorithm"]\\

    Output:\\
    \{\{
      "Part-of-Speech Tagging": 1,
      "NER": 1,
      "Graph Algorithm": 0
    \}\}\\

    You should only output the dict in str format, donnot output analysis.\\\\
    Now evaluate:\\
    Raw Tags: \{raw\_tags\}\\
    Generated Tags: \{generate\_tags\}
\end{tcolorbox}
\caption{Instruction template for Precision Eval.}
\label{tab:precision_eval}
\end{table*}

\begin{table*}
\begin{tcolorbox}[colback=gray!10,
                  colframe=black,
                  width=1\textwidth,
                  arc=1 mm, 
                  boxrule=0.5pt,
                  title=Instruction template for Precision Eval,
                  fontupper=\small
                 ]
You are given a list of generated tags.\\\\
    Your task is to determine for each tag whether it is a **fine-grained** knowledge concept.\\

    Criteria:\\
    - Fine-grained tags are specific, well-defined concepts or techniques.\\
    - Tags that are broad should be marked as not fine-grained.\\
    - A moderate level of granularity is also acceptable — for example, ‘solving quadratic equations’ is considered fine-grained.\\
    - For example, "Quick Sort" is fine-grained, but "Sorting Algorithms" is not.\\

    \#\#\# Output format:\\
    Return a dictionary where each key is a generated tag, and the value is `1` if it is fine-grained, or `0` if not.\\

    \#\#\# Example:\\
    Generated Tags: ["Quick Sort", "Sorting Algorithms", "Graph Traversal"]\\

    Output:\\
    \{\{
      "Quick Sort": 1,
      "Sorting Algorithms": 0,
      "Graph Traversal": 0
    \}\}\\

    Only output the dictionary in string format without any explanation.\\

    Now evaluate:\\
    Generated Tags: \{generate\_tags\}
\end{tcolorbox}
\caption{Instruction template for Fine-grained Eval.}
\label{tab:finegrained_eval}
\end{table*}

\subsection{TreeEval}
We further prompt the LLM to evaluate the constructed Tag Tree from Instance-Level (Table~\ref{tab:instance_level}) and Cluster-Level (Table~\ref{tab:cluster_level}) dimensions.
\begin{table*}
\begin{tcolorbox}[colback=gray!10,
                  colframe=black,
                  width=1\textwidth,
                  arc=1 mm, 
                  boxrule=0.5pt,
                  title=Instruction template for Instance-level Eval,
                  fontupper=\small
                 ]
You are an expert taxonomist. Your task is to evaluate hierarchical relationships between  concepts. Please evaluate the relationship between the Parent and Child:\\

Parent Topic: \{name\}\\
Child Topic: \{cluster\_item\}\\

Criteria:\\
- Return 1 if the Child Topic is logically a sub-topic, component, or relevant instance of the Parent Topic.\\
- Return 0 if the Child Topic is unrelated or clearly incorrect under this parent.\\

Output format: Return ONLY the single digit (0 or 1). \\
\end{tcolorbox}
\caption{Instruction template for Instance-Level Eval.}
\label{tab:instance_level}
\end{table*}

\begin{table*}
\begin{tcolorbox}[colback=gray!10,
                  colframe=black,
                  width=1\textwidth,
                  arc=1 mm, 
                  boxrule=0.5pt,
                  title=Instruction template for Cluster-Level Eval,
                  fontupper=\small
                 ]
You are an expert taxonomist and data evaluator. Your task is to evaluate if a given Topic Name  accurately reflects a list of 'Contained Clusters'.\\
Please evaluate the following:\\

Topic Name: \{name\}\\
Contained Clusters: \{clusters\_str\}\\

Criteria:\\
- Return 1 if the Topic Name is a high-quality, accurate summary that 
encompasses the Contained Clusters.\\
- Return 0 if the Topic Name is irrelevant, misleading, or fails to cover the main themes of the clusters.\\

Output format: Return ONLY the single digit (0 or 1). 
\end{tcolorbox}
\caption{Instruction template for Cluster-Level Eval.}
\label{tab:cluster_level}
\end{table*}

\end{document}